\documentclass[a4paper,12pt]{article}

\usepackage{cite}
\usepackage{amsmath,amssymb,amsfonts,color}
\usepackage{algorithmic}
\usepackage{graphicx}
\usepackage{textcomp}
\usepackage{xcolor}
\usepackage{subcaption} 
\usepackage[ruled,vlined]{algorithm2e}

\def\BibTeX{{\rm B\kern-.05em{\sc i\kern-.025em b}\kern-.08em
    T\kern-.1667em\lower.7ex\hbox{E}\kern-.125emX}}
\begin{document}
%
\title{A Comparison of Methods for Neural Network Aggregation}
%
%
%

\author{John Pomerat
        Aviv Segev
\thanks{Department
of Computer Science, University of South Alabama, Mobile,
AL, 36688 USA e-mail: segev@southalabama.edu.
}
}

%
%

\markboth{Journal of \LaTeX\ Class Files,~Vol.~6, No.~1, January~2007}%
{Shell \MakeLowercase{\textit{et al.}}: Bare Demo of IEEEtran.cls for Journals}
%



\date{}
\maketitle
\thispagestyle{empty}

\begin{abstract}
Deep learning has been successful in the theoretical aspect. For deep learning to succeed in industry, we need to have algorithms capable of handling many inconsistencies appearing in real data. These inconsistencies can have large effects on the implementation of a deep learning algorithm. Artificial Intelligence is currently changing the medical industry. However, receiving authorization to use medical data for training machine learning algorithms is a huge hurdle. A possible solution is sharing the data without sharing the patient information. We propose a multi-party computation protocol for the deep learning algorithm. The protocol enables to conserve both the privacy and the security of the training data. Three approaches of neural networks assembly are analyzed: transfer learning, average ensemble learning, and series network learning. The results are compared to approaches based on data-sharing in different experiments. We analyze the security issues of the proposed protocol. Although the analysis is based on medical data, the results of multi-party computation of machine learning training are theoretical and can be implemented in multiple research areas.
\end{abstract}


\providecommand{\keywords}[1]
{
  \small	
  \textbf{\textit{Neural network aggregation, Multi-party computation, Transfer learning, Average ensemble learning}} #1
}

%

\section{Introduction}
In recent years, the theoretical progress of machine learning promises to revolutionize many domains of industry, from manufacturing\cite{1}, through healthcare \cite{2} and transportation\cite{19}, to education\cite{20}. Although there have been many successful implementations of learning algorithms, much of the progress in machine learning remains theoretical \cite{3}. One reason for the lack of implementation, particularly in the healthcare domain, is the practicality, resilience, and security of learning algorithms \cite{2,4}. A staple of machine learning is data; as such, its shape, organization, quantity, and quality must all be carefully considered for many real-world implementations \cite{5}. As the need for healthcare datasets rises, data-sharing \cite{5} has been suggested as a strategy to get data for healthcare models. In data-sharing, hospitals reformat data into an agreed upon structure and anonymize contents so as not to expose confidential patient data. We propose an alternative to data-sharing using secure multi-party computation (MPC). Multi-party computation is a branch of cryptography concerned with calculating functions on private, user-held data. One motivating example considers two people who wish to determine which of them has a higher salary without either party exposing their salary to one-another. With MPC, there is an algorithm capable of solving this problem and other, similar, problems. We propose a protocol for training neural networks on private datasets then combining the neural networks such that private data is not exposed, and the model's final performance is comparable to a model trained on the combined private datasets. This paper considers three methods of neural network aggregation to combine networks trained on distinct datasets sharing an underlying function. For all three of the methods, underlying network architecture, datasets were kept constant. Additionally, hyperparameters, including activation functions, optimizer, and batch size were held constant. Performance was measured by mean square error, which was recorded to compare the three methods. The three methods are transfer learning, average ensemble learning, and series network learning.

This paper will explore these methods of neural network aggregation in depth.
\section{Related Work}
\subsection{Security}
In the healthcare domain, the importance of maintaining data privacy is clear. As such, sensitive data should be anonymized as much as possible to prevent any kind of data leakage. There are a number of attacks against learning algorithms \cite{6,7,8,9}. In 2015, Goodfellow et al. proposed adversarial attacks as a security vulnerability of neural networks \cite{10}. Since then, there has been more research into the security of neural networks and more attack vectors have been discovered \cite{6,7,8,9}. Our problem, as we have defined it, is not vulnerable to black box adversarial attacks. One attack vector that our system is vulnerable to is training code provided by a malicious adversary \cite{11}. To protect against this attack, the code which specific parties implement should be open-sourced and independently reviewed. Additionally, there is an attack vector for generative models\cite{12} which should be considered for some implementations with generative models but is beyond the scope of this paper. The primary attack vector of concern is the membership inference attack \cite{18}. The membership inference attack is a blackbox attack vector for a trained neural network classifier. The attack is an algorithm to statistically determine from a trained neural network whether an input tuple is a member of the private training set or not\cite{18}. To protect against the membership inference attack, models should avoid overfitting. Additionally, adding regularization, prediction vector restrictions, and increasing the entropy of the prediction vector have value in preventing membership inference attacks \cite{18}.

\subsection{Transfer Learning}
A well known method of neural network aggregation is transfer learning. Recently, transfer learning has been shown to be useful and extremely versatile, particularly with reinforcement learning and deep neural network models\cite{22,23,16,24}. Additionally, transfer learning is also more versatile than some of the methods explored in this paper since it is capable of working with a wider variety of learning algorithms including convolutional neural networks\cite{22}. Furthermore, transfer learning has been shown to work well with time series predictions and recurrent neural networks \cite{25}. In addition, research on transfer learning in the context of the healthcare domain already shows promise \cite{26,27,28}. One paper by Gupta et al.\cite{26} leveraged transfer learning to generalize models in the healthcare domain to similar tasks in the same domain. Similarly, a result by Chen et al.\cite{27} with wearable technology used transfer learning on time series health data to improve performance and increase personalization of the FedHealth model. The results for transfer learning show promise for the viability of neural network aggregation for deep learning in the healthcare domain.\newline   Previous work in transfer learning shows great promise for neural network aggregation as an alternative to data-sharing in big data healthcare applications. These results, combined with some of the research conducted in security and multi-party computation act as the motivating examples for this paper.

\section{Neural Network Aggregation}
\subsection{Problem Statement}
The setup goes as follows. Let $D_1, D_2, \cdots, D_k$ be subsets of $\mathbb{R}^n$ represented as datasets. Then, let $G: \mathbb{R}^n \to \mathbb{R}^m$ be a differentiable function represented as a multilayer perceptron with parameters $\theta _g$. We are concerned with methods of producing $\theta _g$ from the $D_i$ such that the loss of $G$ is comparable to obtaining the $\theta _g$ from $\bigcap_{i=1}^k D_{i}$ and it is not computationally feasible to extract information about the members of the $D_i$ from $G$. This process, of training a neural network from multiple, disjoint datasets is called neural network aggregation. The three methods of neural network aggregation are series network learning, average ensemble learning, and transfer learning.

\subsection{Series Network Learning}
\begin{figure}[h]
    \centering
    \includegraphics[width=1\linewidth]{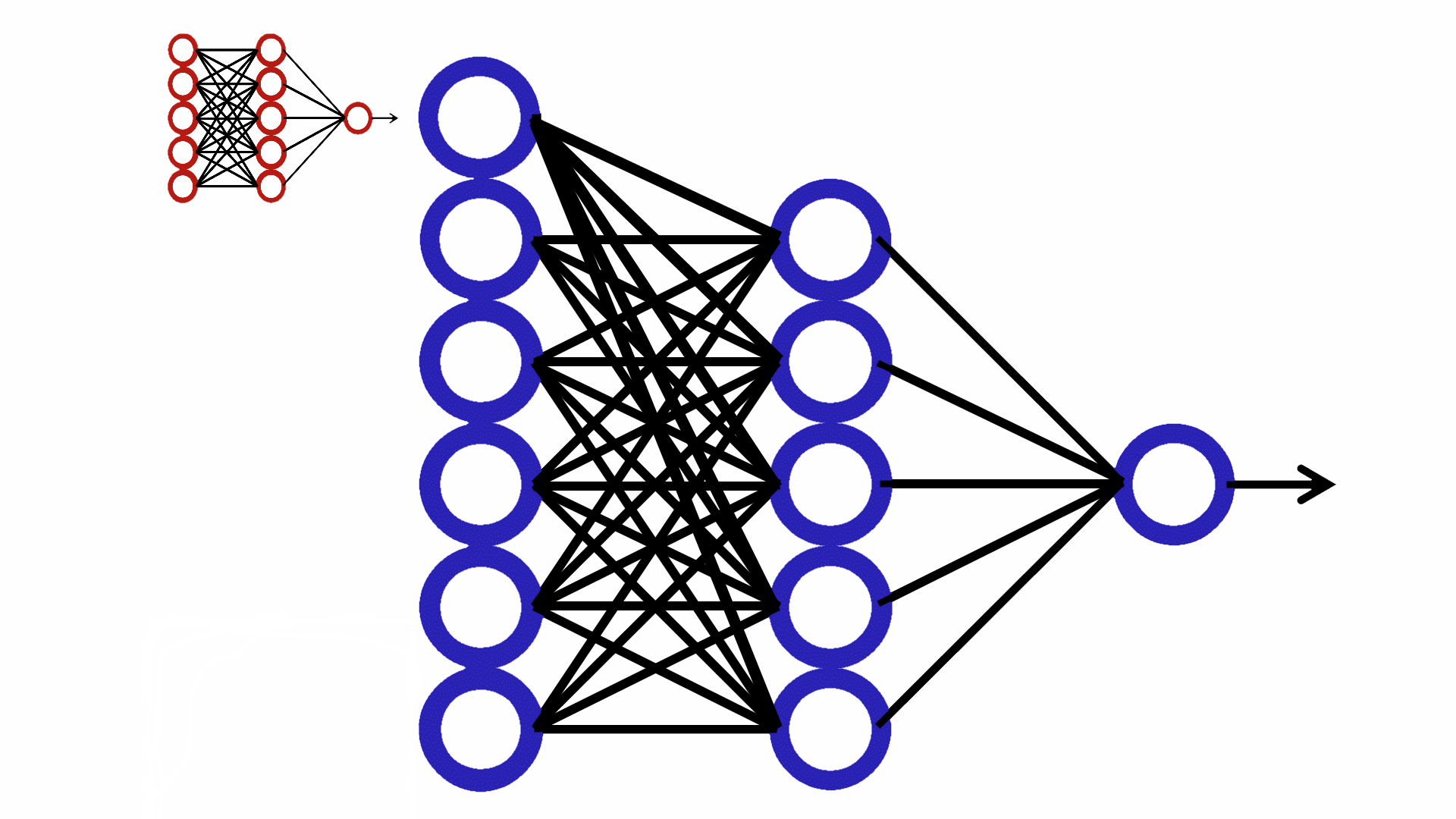}
    \caption{Neural network as an input to assist training a second network.}
    \label{fig:NN1}
\end{figure}

The first method, called series network learning, functions by training a neural network with a pretrained ``expert" neural network as an additional input. For our experiment we consider a single neural network trained on the first dataset. The neural network's performance on the testing set is recorded. The network then generates a prediction for every entry in the second dataset, a new neural network is then created for the second dataset with the prediction array as an additional input. The neural network is then trained on the second dataset and the mean square error is recorded. Intuitively, the second neural network will likely have an improvement in mean square error as the network will ``learn" when to trust the first network's prediction and when to instead use its own calculations, Fig. \ref{fig:NN1}. 

\begin{algorithm}[H]
\SetAlgoLined
 \For{For all parties except the last}{
  train network on parties data\;
 }
 take network and append each output of the trained networks as a new input neuron then train resulting network on the final parties data;
 \caption{Series Networks}
\end{algorithm}

\subsection{Average Ensemble Learning}
\begin{figure}[h]
    \centering
    \includegraphics[width=1\linewidth]{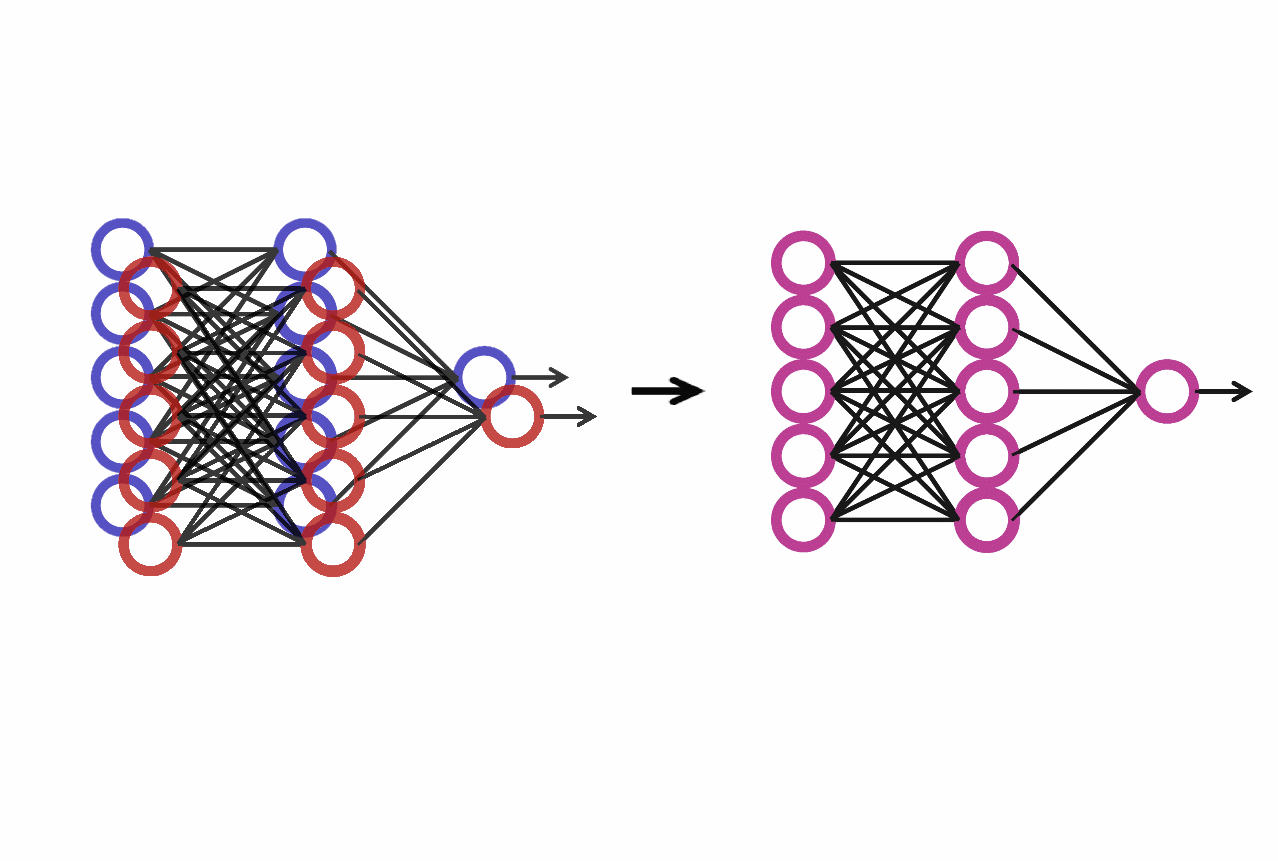}
    \caption{Averaging the corresponding weights and biases of a neural network}
    \label{fig:NN2}
\end{figure}

The second method considers two neural networks, $N_1, N_2$ of the exact same architecture, with the same activation functions, optimizer, number of hidden layers, and number of neurons. Each of the networks are then trained on different datasets of identical structure and mean square error on the testing set is recorded. Then, the two neural networks are combined to form a third network of the exact same structure $N_3$ (Fig. \ref{fig:NN2}). The weights and biases of $N_3$ are the average of the corresponding weights and biases in $N_1$ and $N_2$. More specifically, if $n$ is the total number of weights and biases in $N_3$, and $N_3(i)$ refers to the $i$-th weight or bias in $N_3$, then for all $0<i \le n$,
$$N_3(i) = \frac{N_1(i) + N_2(i)}{2}$$
$N_3$ is then measured on the testing set and its performance is compared to the performance of both $N_1$ and $N_2$. In addition to a pure average, other strategies are considered. Initially, a weighted average may be performed with weights proportional to the size of the dataset to guarantee that a model trained on significantly more data is not treated the same as a model trained on a much smaller set of data. Another option is to use a weighted average not only with the size of the dataset, but also the ratio of positives and negatives for disease prediction cases. This is done to ensure that a larger dataset, which is not highly informative, will not overpower a smaller dataset containing more information.

\begin{algorithm}[h]
\SetAlgoLined
 \For{For all parties}{
  train identical network on parties data\;
 }
 initialize new model identical to the others;
 
 \For{every weight and bias in the network}{
    \For{every trained network}{
        sum values of corresponding weight or bias;
    }
    weight or bias in new network is that sum divided by the number of parties;
 }
 \caption{Average Ensemble Learning}
\end{algorithm}

\subsection{Transfer Learning}
The third method is transfer learning. Instead of combining two neural networks, transfer learning functions by training on additional datasets with a single neural network without weight reinitialization\cite{16, 17}. Our experiment considers a single neural network with randomly initialized weights trained on the first dataset. The mean squared error on the testing set is recorded. Then, the neural network is trained on the second dataset without reinitializing the weights and the mean squared error is recorded again. This process is then repeated by training on the second dataset and then the first. With mean square error being recorded throughout.

\section{Experiments}
To compare the proposed methods of neural network aggregation, we ran two experiments, one with artificially generated polynomial data, and the other on the University of Wisconsin Madison Hospital's breast cancer dataset\cite{21}. The motivation of these tests was to get an initial performance comparison between the proposed methods and a neural network trained on all of the data simultaneously representing data-sharing.
\subsection{Data}
\begin{figure}[h]
    \centering
    \includegraphics[width=.9\linewidth]{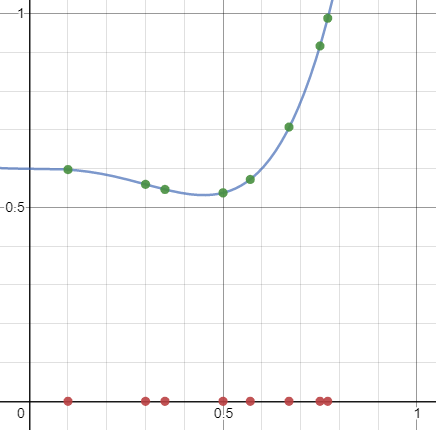}
    \caption{Random x values (red) and calculated y values (green) on the generated polynomial function (blue).}
    \label{fig:exampledata}
\end{figure}

The neural networks in this paper were trained on both real and artificially generated data. The artificially generated data was created as follows. A random normal distribution was employed to create 2 dimensional arrays populated with random rational numbers in a specified range. The rows of the array consisted of 7 random rational numbers representing data features. Multiple datasets were created for the experiment. Arrays of size 3200, 1600, 800, and 400 were created. After the arrays were generated, a multivariate polynomial of degree $n$ under lexicographic term ordering was created with coefficients randomly chosen from a normal distribution.
\begin{equation}
    f(x_1,x_2...x_7)
\end{equation} Next, for each set of $7$ values in the generated data, $\gamma$, $f(\gamma)$ was calculated by plugging the values from the generated data into the polynomial Fig. \ref{fig:exampledata} illustrates this in $2$ dimensions as opposed to $7$.

After $f(\gamma)$ has been calculated for all the tuples in each array, the values were combined with the generated data to form a dataset such that each row contains 8 values, 7 random rational numbers, and the calculated y-value according to the generated function. Thus, the networks in the experiments will be trained on the 7 rational numbers to learn the underlying polynomial function. These datasets were then divided into two training sets and a testing set containing 80\% and 20\% of the entries respectively. The training set was then divided again into two training sets of equal size.

The real data used in this paper comes from the University of Wisconsin Madison Hospital's breast cancer dataset\cite{21}. This dataset was also divided into two equally sized training sets and a single testing set. The breast cancer dataset contains 569 rows and 32 features. This was split into two training sets each with 256 training examples and a testing set with 57 examples. The features in the breast cancer data describe tumors. Some of the features include clump thickness, uniformity of cell size and shape, marginal adhesion, and others. Furthermore, each of these features was recorded in three different ways in the data. For each feature, an average, a low, and a high value were all available in the data. The data preprocessing used consisted of normalization and minor feature manipulations to get the data in the right shape to form proper training and testing sets.

\subsection{Regression}
For this experiment, the regression data (as defined above) was taken, then split into a training set and a testing set with $80\%$ of the examples for training and $20\%$ of the examples for testing. A neural network was trained on the training set, then loss on testing set was recorded. Since all the data was in one place, the resulting model represents a network trained on a dataset created through data-sharing. Then, the training set was split into two smaller training sets of equal size. Then, we perform each of the three methods to train a neural network from the split datasets recording loss for each. The neural network architecture was chosen to best fit the data and was kept constant while the test was repeated many times under different conditions. The conditions varied epochs from 10 to 200, noise in the regression data from a logarithmic shift with coefficients varying between 1 and 3, size of the datasets from 400 to 32,000, and polynomial degree of the underlying dataset from 2 to 5. The average loss for all methods, including the loss from the data-sharing model represented as ``None", across all tests can be seen in below in Fig. \ref{fig:graph1}. Additionally, average loss for tests with varying degrees of added noise can be found in Table \ref{tab:my_label2}. The added noise in the data is given by $$y = f(x_1,x_2,...x_7) + ndr$$ where $x_1, x_2 ... x_7$ is a data point, $n$ a chosen noise value, $r$ is a random real number selected from a random normal distribution between -2 and 2, and $f$ is a polynomial function with degree $d$.

\begin{figure}[h]
    \centering
    \includegraphics[width=1\linewidth]{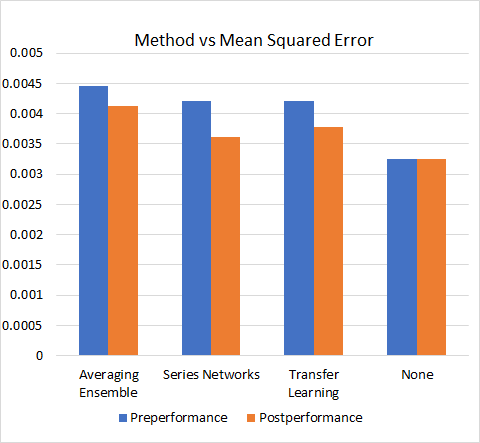}
    \caption{Loss comparison of methods on regression data}
    \label{fig:graph1}
\end{figure}

\begin{table}[h]
    \centering
    \begin{tabular}{c|c|c}
         Method & Average MSE & Noise n \\
             \hline
          & 0.015 & 0  \\
         Average Ensemble & 0.011 & 1 \\
          & 0.011 & 2  \\
          \hline
          & 0.013 & 0  \\
         Series Networks & 0.010 & 1 \\
          & 0.010 & 2  \\
          \hline
          & 0.011 & 0  \\
         Transfer Learning & 0.007 & 1 \\
          & 0.009 & 2 \\
          \hline
          & 0.006 & 0  \\
         None & 0.008 & 1  \\
          & 0.008 & 2  \\
         
    \end{tabular}
    \caption{Loss comparison for methods with added noise}
    \label{tab:my_label2}
\end{table}
Preperformance is the loss measured on the testing set once the model had learned on the first dataset. Similarly, postperformance is the loss measured after the second dataset had been aggregated in. The purpose of this is to see the method converge to the performance of the model obtained through data-sharing by aggregating in multiple datasets. After training, all three methods achieved comparable aggregate performance compared to the model trained on the combined ``shared" data (None). Here, series networks had the best performance of the three methods and also had the greatest performance increase after aggregation.

\subsection{Breast Cancer Classification}
\begin{figure}[h]
    \centering
    \includegraphics[width=1\linewidth]{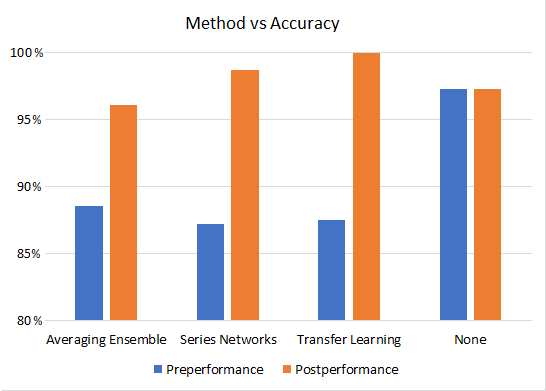}
    \caption{Accuracy comparison of methods on breast cancer data}
    \label{fig:graph2}
\end{figure}

\begin{figure}[h]
    \centering
    \includegraphics[width=1.1\linewidth]{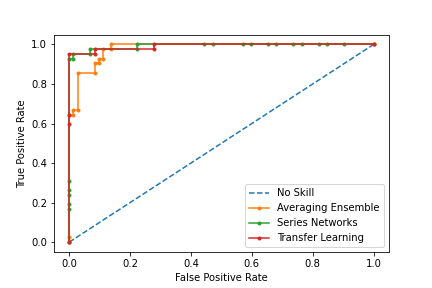}
    \caption{ROC Curve for Breast Cancer Data}
    \label{fig:graph3}
\end{figure}

For this experiment, the goal is to train a classifier to determine whether a tumor is benign or malignant. The breast cancer dataset contains 569 rows and 32 features. Similarly to the regression experiment, the data was split, a neural network architecture was configured for the data, then accuracy values for the three methods were computed. Additionally, the data-sharing equivalent model was trained on the data before the training sets were bifurcated and the accuracy was recorded. Tests were repeated with varied hyperparameters, including, batch size, epochs, and number of neurons. The accuracy values of the test can be found in Fig. \ref{fig:graph2}, the ROC curve for the test is in Fig. \ref{fig:graph3}, and precision, recall, and F1 scores are in Table \ref{tab:my_label}.

Here, all of the methods performed better than the equivalent model obtained through data-sharing. Additionally, this example also provides evidence for the viability of our method in the healthcare domain.

\begin{table}[h]
    \centering
    \begin{tabular}{c|c|c|c}
         Method & Precision & Recall & F1 Score \\
             \hline
         Average Ensemble & 0.76 & 1.00 & 0.87 \\
         Series Networks & 1.00 & 0.93 & 0.96 \\
         Transfer Learning & 1.00 & 0.93 & 0.96
    \end{tabular}
    \caption{Metrics for compared methods}
    \label{tab:my_label}
\end{table}

From the accuracy graph (Fig. \ref{fig:graph2}) and the ROC curve (Fig. \ref{fig:graph3}), transfer learning and series networks performed the best, outperforming training on the combined dataset. This is likely due to the fact that with smaller dataset size, training on smaller subsets of the data grants more generalization.

\section{Conclusion}
In order for neural network aggregation to be fully recognized as a stronger alternative to data-sharing, more tests need to be run. Additionally, future work should examine the scaling of the proposed model, examining for model convergence as the number of disjoint datasets increases. If transfer learning or series network learning is able to converge to the same model acquired through data-sharing by distributing training across many datasets, then the method would be viable. Furthermore, more studies need to be conducted on membership inference attacks to lower the security concerns. Since the membership inference attack is strong against overfit models, it would be interesting to see what the end behavior of series networks or transfer learning after training on many datasets. Despite this, both transfer learning and series network learning seem to be promising methods for distributing training on private datasets. Thus, while more tests need to be conducted to prove the viability of neural network aggregation, this paper establishes an initial claim for neural network aggregation as a functional alternative to data-sharing.

\end{document}